\documentclass[twocolumn, letterpaper]{IEEEAerospaceCLS}  % only supports two-column, letterpaper format

\usepackage{cite} 
\usepackage[]{graphicx}    % We use this package in this document
\usepackage[]{amsmath}
\usepackage{textcomp}
\usepackage{xcolor}
\usepackage{caption}
\usepackage{subcaption}
\usepackage{lipsum}
\usepackage{mathtools}
\usepackage{cuted}
\usepackage[]{xcolor, soul}
\sethlcolor{yellow}
\usepackage{multirow}
\usepackage{booktabs}

\newcommand{\ignore}[1]{}  % {} empty inside = %% comment
\def\BibTeX{{\rm B\kern-.05em{\sc i\kern-.025em b}\kern-.08em
    T\kern-.1667em\lower.7ex\hbox{E}\kern-.125emX}}
    
\begin{document}
\title{Testing Spacecraft Formation Flying with Crazyflie Drones as Satellite Surrogates\thanks{Approved for public release; distribution is unlimited. AFRL-2023-4302}}

\author{
Arturo de la Barcena*\\
MECE Department\\
University of Houston\\
Houston, TX 77204\\
adelabarcena@uh.edu
\and
Collin Rhodes*\\ 
MECE Department\\
University of Houston\\
Houston, TX 77204\\
cjrhodes@cougarnet.uh.edu
\and
John McCarroll\\
Matrix Research\\
3844 Research Blvd\\
Dayton, OH 45430\\
john.mccarroll.ext@afresearchlab.com
\and
Marzia Cescon\\
MECE Department\\
University of Houston\\
Houston, TX 77204\\
mcescon2@central.uh.edu
\and
Kerianne L. Hobbs\\ 
Air Force Research Laboratory\\
2241 Avionics Circle\\
Wright-Patterson AFB, OH, 45433\\
kerianne.hobbs@us.af.mil
%%%% IMPORTANT: Use the correct copyright information--IEEE, Crown, or U.S. government. %%%%%
\thanks{\footnotesize 979-8-3503-0462-6/24/$\$31.00$ \copyright2024 IEEE\\ Equal contribution*}              % This creates the copyright info that is the correct 2024 data.
%\thanks{{U.S. Government work not protected by U.S. copyright}}         % Use this copyright notice only if you are employed by the U.S. Government.
%\thanks{{979-8-3503-0462-6/24/$\$31.00$ \copyright2024 Crown}}          % Use this copyright notice only if you are employed by a crown government (e.g., Canada, UK, Australia).
%\thanks{{979-8-3503-0462-6/24/$\$31.00$ \copyright2024 European Union}}    % Use this copyright notice is you are employed by the European Union.
}

\maketitle
\thispagestyle{plain}
\pagestyle{plain}

%%%%%%%%%%%%%%%%%%%%%%%%%%%%%%%%%%%%%%%%%%%
\begin{abstract}
As the space domain becomes increasingly congested, autonomy is proposed as one approach to enable small numbers of human ground operators to manage large constellations of satellites and tackle more complex missions such as on-orbit or in-space servicing, assembly, and manufacturing. One of the biggest challenges in developing novel spacecraft autonomy is mechanisms to test and evaluate their performance. Testing spacecraft autonomy on-orbit can be high risk and prohibitively expensive. An alternative method is to test autonomy terrestrially using satellite surrogates such as attitude test beds on air bearings or drones for translational motion visualization. Against this background, this work develops an approach to evaluate autonomous spacecraft behavior using a surrogate platform, namely a micro-quadcopter drone developed by the Bitcraze team, the Crazyflie 2.1. The Crazyflie drones are increasingly becoming ubiquitous in flight testing labs because they are affordable, open source, readily available, and include expansion decks which allow for features such as positioning systems, distance and/or motion sensors, wireless charging, and AI capabilities. In this paper, models of Crazyflie drones are used to simulate the relative motion dynamics of spacecraft under linearized Clohessy-Wiltshire dynamics in elliptical natural motion trajectories, in pre-generated docking trajectories, and via trajectories output by neural network control systems.
\end{abstract}

%%%%%%%%%%%%%%%%%%%%%%%%%%%%%%%%%%%%%%%%%%%
\tableofcontents

%%%%%%%%%%%%%%%%%%%%%%%%%%%%%%%%%%%%%%%%%%%
\section{Introduction}
The control of large constellations of satellites has seen increased attention in a variety of fields including Earth sciences and astrophysics. Commercially it has been considered for missions including space and Earth surveillance, weather forecasting, data transmission \cite{bandyopadhyay2015review}, and, most recently, on-orbit/in-space servicing, assembly, and manufacturing \cite{piskorz2018orbit}. As mission complexity advances in these fields, autonomy has been introduced as a method of enabling a small number of human ground operators to manage these large satellite constellations.  Autonomous satellite formation flight has been the subject of study for many years with advancements in collision avoidance \cite{slater2006collision}, optimal control through trajectory tracking \cite{petit2001constrained}, and control through reinforcement learning \cite{campbell2017deep} \cite{lei2022deep} occupying a large part of related literature over the past twenty years. Spacecraft autonomy testing, however, can be prohibitively difficult and expensive when done on orbit. It follows that the development of simple, scalable, inexpensive testing is highly desirable, with the use of terrestrial satellite surrogates being a promising method. To date, several research groups have considered using terrestrial, robotic surrogates. For instance, the authors in  \cite{pickem2016safe} and \cite{pickem2017robotarium} created a remote access swarm robotarium as a robotics educational tool, and a two-dimensional representation of deep reinforcement learning controlled spacecraft was modeled on ground-based robots in \cite{hovell2021deep}, while \cite{Lippay2024Emulation} describes the Local Intelligent Networked Collaborative Satellites (LINCS) Lab, which employs linear quadratic regulation (LQR) control on aerial vehicles to follow satellite dynamics. This work develops an approach to simulate satellite relative motion dynamics under the linearized Clohessy-Wiltshire equations, which describe the three-dimensional motion of a chaser spacecraft with respect to a target spacecraft in the Hill's reference frame, using a surrogate platform capable of representing translational motion in three dimensions. The surrogate of choice in this paper is the Bitcraze Crazyflie 2.1 micro-quadcopter platform~\cite{crazyflie_21}. These drones are of interest because they are open source, inexpensive, readily available, and include up to two expansion decks which allow for features such as positioning systems, distance and/or motion sensors, wireless charging, and AI capabilities. Additionally, at 27 grams \cite{giernacki2017crazyflie}, their small size and affordability facilitates testing cooperative control algorithms on many vehicles in smaller lab spaces, further reducing the requirements for testing. In this work, the \texttt{gym-pybullet-drones} simulation platform~\cite{panerati2021learning}
%, with demonstrated sim-to-real transition,
is leveraged as a virtual environment to simulate the Crazyflie 2.1 drones. Natural motion and controlled trajectories are determined in the space scale, then scaled and simulated in this virtual environment.

% \textcolor{red}{Find sources for the cost and risk on in orbit testing to include discussion about the benefits of terrestrial testing}
%%%% If spacecraft autonomy controls are implemented, they NEED to be mentioned in the Intro %%%%

%%%%%%%%%%%%%%%%%%%%%%%%%%%%%%%%%%%%%%%%%%%
\section{Background}\label{Sec:background}
This section introduces how crazyflie drones can be used as a satellite surrogate test bed, the linearized relative motion spacecraft dynamics, and the dynamics and control of the crazyflie drones.

\subsection{Crazyflies as a Satellite Surrogate Test Bed}
The terrestrial surrogate simulated in this paper is the Bitcraze Crazyflie 2.1 micro-quadcopter running in the \texttt{gym-pybullet-drones} gym environment \cite{panerati2021learning}. The \texttt{gym-pybullet-drones} environment is an open-source, gym-style environment based on Google’s Bullet Physics engine \cite{himanshu2022waypoint} using its Python binding, Pybullet, and is one of the first general purpose multi-agent Gym environments for quadcopters. The physics used to model the quadcopter is based on system identification performed on the actual Crazyflie 2.1 platform \cite{panerati2021learning} and a separate study done on different physical phenomena including drag, the ground effect, and the downwash effect \cite{forster2015system}. These inexpensive micro-quadcopters have become commonplace in lab environments and provide an easy-to-use platform for satellite formation visualization with the Crazyflies serving as scaled satellite surrogates.

\subsection{Spacecraft Linearized Clohessy-Wiltshire Dynamics}
\begin{figure}[htb!]
    \centering
    \includegraphics[width = .49\textwidth]{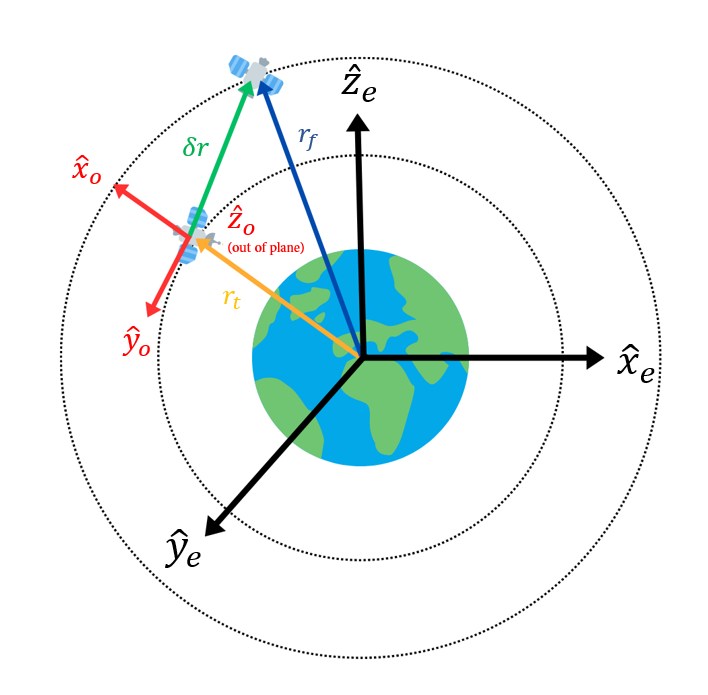}
    \caption{Earth-Centered Inertial Frame (Black) and the Non-inertial Hill's Frame (Red). The Hill's frame x-axis points from the center of the Earth to the chief satellite's center of mass, the z-axis points in the same direction as the orbital angular moment (out of the page), and the y-axis completes the orthogonal frame.}
    \label{fig:HillsFrame}
\end{figure}

%A space dynamics simulator was used to create a drone trajectory within the \texttt{gym-pybullet-drones} environment. This simulator solves the Clohessy-Wiltshire (CW) equations \cite{clohessy1960terminal}, which describe the relative motion of one spacecraft to another in the non-inertial Hill's reference frame \cite{hill1878researches}, depicted in Fig. \ref{fig:HillsFrame}.
A space dynamics simulator was used to create spacecraft trajectories solving the Clohessy-Wiltshire (CW) equations \cite{clohessy1960terminal}, which describe the motion of one spacecraft relative to another in the non-inertial Hill's reference frame \cite{hill1878researches}, depicted in Fig. \ref{fig:HillsFrame}.

A linearization of the relative motion dynamics between the deputy and chief spacecraft is given by

\begin{equation}
{\bf \dot{x}} = {A \bf{x} + B \bf{u}}
\label{eqn: CWEquation}
\end{equation}
where,
\begin{equation}
A =\ 
\begin{bmatrix}
    0 & 0 & 0 & 1 & 0 & 0\\
    0 & 0 & 0 & 0 & 1 & 0\\
    0 & 0 & 0 & 0 & 0 & 1\\
    3n^2 & 0 & 0 & 0 & 2n & 0\\
    0 & 0 & 0 & -2n & 0 & 0\\
    0 & 0 & -n^2 & 0 & 0 & 0\\
\end{bmatrix}
\end{equation}
and
\begin{equation}
B =\
\begin{bmatrix}
    0 & 0 & 0\\
    0 & 0 & 0\\
    0 & 0 & 0\\
    1/m & 0 & 0\\
    0 & 1/m & 0\\
    0 & 0 & 1/m\\
\end{bmatrix}
\end{equation}
with $n$ representing the satellite's mean orbital motion (an inverse relationship of the orbital period) and $m$ the satellite's mass. For Earth-based applications, $n$ = 0.001027 [rad/s]. The states in this equation are given by the relative positions and velocities in the $x$, $y$, and $z$ directions and the control is given by the thrust forces in these directions. This linearization is an approximation of the exact equations of relative motion that assumes the distance between the two satellites is much less than the distance between either satellite and the Earth, and that the chief satellite follows a perfectly circular orbit. Alternatively, these equations can be represented in discrete time \cite{nakka2022information} as using $A$ in Eq. \eqref{eq:Amatrix} and $B$ defined in Eq. \eqref{eq:Bmatrix}.
\begin{figure*}[htb!]
\begin{equation}\label{eq:Amatrix}
A_k =\
\begin{bmatrix}
    4-3\cos(nt) & 0 & 0 & \frac{1}{n}\sin(nt) & \frac{2}{n}(1-\cos(nt)) & 0\\
    6(\sin(nt)-nt) & 1 & 0 & \frac{-2}{n}(1-\cos(nt)) & \frac{1}{n}(4\sin(nt)-3nt) & 0\\
    0 & 0 & \cos(nt) & 0 & 0 & \frac{1}{n}\sin(nt)\\
    3n\sin(nt) & 0 & 0 & \cos(nt) & 2\sin(nt) & 0\\
    -6n(1-\cos(nt)) & 0 & 0 & -2\sin(nt) & 4\cos(nt)-3 & 0\\
    0 & 0 & -n\sin(nt) & 0 & 0 & \cos(nt)\\
\end{bmatrix}
\end{equation}
\end{figure*}

\begin{figure*}
\begin{equation}\label{eq:Bmatrix}
B_k =\
\begin{bmatrix}
    \frac{1}{n}(\cos(nt)-1) & \frac{2}{n}(t+\frac{1}{n}\sin(nt)) & 0\\
    \frac{-2}{n}(t-\frac{1}{n}\sin(nt)) & \frac{1}{n}(\frac{-4}{n}(\cos(nt)-1)-(3n/2)t^2) & 0\\
    0 & 0 & \frac{-1}{n^2}(\cos(nt)-1)\\
    \frac{1}{n}\sin(nt) & \frac{-2}{n}(\cos(nt)-1) & 0\\
    \frac{2}{n}(\cos(nt)-1) & \frac{4}{n}(\sin(nt)-3t) & 0\\
    0 & 0 & \frac{1}{n}\sin(nt)\\
\end{bmatrix}
\end{equation}
\end{figure*}
To create a Natural Motion Trajectory (NMT), the control law $u$ must equal the null vector (there is no control input). The trajectory is dependant on the initial conditions (starting positions and velocities), where a relative elliptical orbit can only be ensured if

\begin{subequations}
    \begin{equation}
        \centering
        \dot{y_0} = -2nx_0
            \label{eqn: Con1}
    \end{equation}
    
    %and
    
    \begin{equation}
        \centering
        y_0 = \frac{2\dot{x_0}}{n}
        \label{eqn:Con2}
    \end{equation}
\end{subequations}

These conditions arise from the closed-form solutions to the Clohessy-Wiltshire equations where the condition in Eq. \eqref{eqn: Con1} ensures zero linear drift with time and Eq. \eqref{eqn:Con2} ensures that there is not a constant offset to the time-dependant solution, meaning that the relative motion is centered about the target spacecraft. It should be noted that the motion out of the orbital plane, in the $z$ direction, is completely decoupled from motion in the other two directions meaning that while an ellipse can be assured, the plane of that ellipse with respect to the orbital plane can be changed. Motion in the $x-y$ plane of the CW equations is called \textit{in-plane} because it is in the orbital plane of the chief satellite, while any motion in the $z$ direction is called \textit{out-of-plane}.

\subsection{Crazyflie dynamics and control model}
\begin{figure}[htb!]
    \centering
    \includegraphics[width = .3\textwidth]{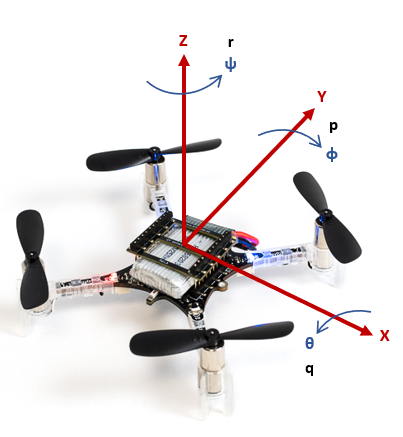}
    \caption{Noninertial Drone Body Fixed Frame.}
    \label{fig:DFrame}
\end{figure}

The state space representation of the linearized dynamics of the Crazyflie 2.1 used in this work was provided in \cite{luis2016design} and with reference to Fig. \ref{fig:DFrame} is described by four decoupled subsystems as follows:
\begin{subequations}
\begin{itemize}
    \item[Vertical Subsystem]
    \begin{equation}
        \begin{bmatrix}
            \Delta\dot{w}\\
            \Delta\dot{z}\\
        \end{bmatrix}
    =
        \begin{bmatrix}
            0 & 0\\
            1 & 0 \\
        \end{bmatrix}
        \begin{bmatrix}
            \Delta w\\
            \Delta z\\
        \end{bmatrix}
    +
        \begin{bmatrix}
            1/m\\
            0\\
        \end{bmatrix}
    \Delta F_z    
    \end{equation}

    \item[Yaw Subsystem]
        \begin{equation}
            \begin{bmatrix}
                \Delta\dot{r}\\
                \Delta\dot{\psi}\\
            \end{bmatrix}
        =
            \begin{bmatrix}
                0 & 0\\
                1 & 0 \\
            \end{bmatrix}
            \begin{bmatrix}
                \Delta r\\
                \Delta \psi\\
            \end{bmatrix}
        +
            \begin{bmatrix}
                1/I_{zz}\\
                0\\
            \end{bmatrix}
        \Delta M_z      
    \end{equation}

    \item[Lateral Subsystem]
        \begin{equation}
            \begin{bmatrix}
                \Delta\dot{p}\\
                \Delta\dot{\phi}\\
                \Delta\dot{v}\\
                \Delta\dot{y}\\
            \end{bmatrix}
        =
            \begin{bmatrix}
                0 & 0 & 0 & 0\\
                1 & 0 & 0 & 0\\
                0 & -g & 0 & 0\\
                0 & 0 & 1 & 0\\
            \end{bmatrix}
            \begin{bmatrix}
                \Delta p\\
                \Delta \phi\\
                \Delta v\\
                \Delta y\\
            \end{bmatrix}
        +
            \begin{bmatrix}
                1/I_{xx}\\
                0\\
                0\\
                0\\
            \end{bmatrix}
        \Delta M_x      
    \end{equation}

       \item[Longitudinal Subsystem]
        \begin{equation}
            \begin{bmatrix}
                \Delta\dot{q}\\
                \Delta\dot{\theta}\\
                \Delta\dot{u}\\
                \Delta\dot{x}\\
            \end{bmatrix}
        =
            \begin{bmatrix}
                0 & 0 & 0 & 0\\
                1 & 0 & 0 & 0\\
                0 & g & 0 & 0\\
                0 & 0 & 1 & 0\\
            \end{bmatrix}
            \begin{bmatrix}
                \Delta q\\
                \Delta \theta\\
                \Delta u\\
                \Delta x\\
            \end{bmatrix}
        +
            \begin{bmatrix}
                1/I_{yy}\\
                0\\
                0\\
                0\\
            \end{bmatrix}
        \Delta M_x      
    \end{equation}
\end{itemize} 
\end{subequations}

%\begin{table}[!h]
%\renewcommand{\arraystretch}{1.7}
%\begin{center}
%\begin{tabular}{ |p{2cm}| |p{2cm}| |p{2cm}|}
%    \hline
%    \multicolumn{3}{|c|}{Drone Variables} \\
%    \hline
%    \multirow{3}{4em}{Moments} & $I_{xx}$ [kg m$^2$] & 1.4e-5\\
%    & $I_{yy}$ [kg m$^2$] & 1.4e-5\\
%    & $I_{zz}$ [kg m$^2$]& 2.17e-5\\
%    \hline
%    Mass & m [kg] & 0.027\\
%    \hline
%\end{tabular}
%\caption{\label{Drone-Variables}Listed variables used in linearized Craziefly 2.1 subsystems.}
%\end{center}
%\end{table}
%
where $ \boldsymbol{x} = [x, y, z] $ represents the drone's linear position relative to a given inertially fixed reference frame, $ \dot{\boldsymbol{x}} = [u, v, w] $ represents the drone's linear velocity, $ \boldsymbol{\theta} = [\theta, \phi, \psi] $ represents the drone's angular position, and $ \boldsymbol{q} = [q, p, r]$ represents the drone's angular velocity. 

\begin{figure}[htb!]
    \centering
    \includegraphics[width = .49\textwidth]{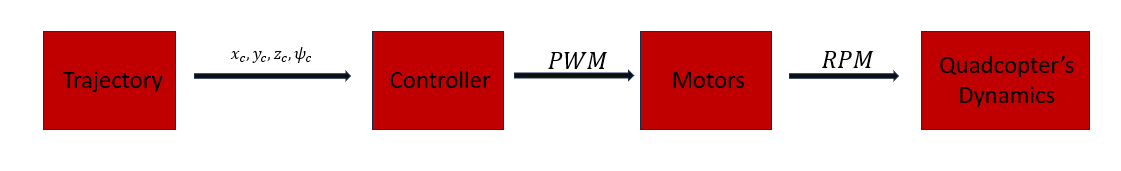}
    \caption{Crazyflie Simulation Control Architecture.}
    \label{fig:SimCom}
\end{figure}

Inputs to these state space realizations are a pulse width modulation (PWM) signal which controls the voltage sent to each direct current (DC) motor, raging from 0 to 65535, which can be related to the motor rotations per minute (RPM) by the following linear relationship \cite{luis2016design}: %are not the angular speeds of each motor in RPM (the angular speeds are linearly related to the input forces and moments), rather 
\begin{equation}
    \centering
    \text{RPM} = 0.2685 * \text{PWM} + 4070.3
\end{equation}
The \texttt{gym-pybullet-drones} gym environment is equipped with a proportional-integral-derivative (PID) controller based on a version of the snap trajectory controller in~\cite{mellinger2011minimum}, also commonly referred to as a Mellinger controller type in some literature \cite{el2023quadcopter}, \cite{javeed2023reinforcement}. This architecture differs from the more common cascaded attitude-position controller by relaxing small pitch ($\theta$) and roll ($\phi$) drone attitude assumptions. The control architecture used in simulation is shown in Fig. \ref{fig:SimCom}.
\begin{table}[b]
\renewcommand{\arraystretch}{1.7}
\begin{center}
\begin{tabular}{ |p{2cm}| |p{2cm}| |p{2cm}|}
    \hline
    \multicolumn{3}{|c|}{Drone Variables} \\
    \hline
    \multirow{3}{4em}{Moments} & $I_{xx}$ [kg m$^2$] & 1.4e-5\\
    & $I_{yy}$ [kg m$^2$] & 1.4e-5\\
    & $I_{zz}$ [kg m$^2$]& 2.17e-5\\
    \hline
    Mass & m [kg] & 0.027\\
    \hline
\end{tabular}
\caption{\label{Drone-Variables}Listed variables used in linearized Craziefly 2.1 subsystems.}
\end{center}
\end{table}

% The first control method found in the Crazyflie 2.1 firmware is a two cascade PID controller, as seen in Fig. \ref{fig:CrazyCS}, which is specified by the manufacturer.

% \begin{figure}[htb!]
%     \centering
%     \includegraphics[width = .49\textwidth]{Figures/cascaded_pid_controller.png}
%     \caption{Crazyflie On-board Control Architecture.}
%     \label{fig:CrazyCS}
% \end{figure}

% This two cascade control structure regulates the angular rate in the inner loop and the attitude in the outer loop with both using PID controllers. The loop is closed by the Crazyflie sensor information gathered by its on-board gyroscope and accelerometer. A snap trajectory controller is, however, also avaliable in the firmware and is comparable to that used in the simulated environment.

%%%%%%%%%%%%%%%%%%%%%%%%%%%%%%%%%%%%%%%%%%%
%\section{Crazyflie Drone Setup}
\section{Experimental Setup}
This section describes the physical laboratory setup and Crazyflie drone swarming and waypoint following capabilities. 

\subsection{Laboratory Setup}
\begin{figure}
    \centering
    \includegraphics[width = .49\textwidth]{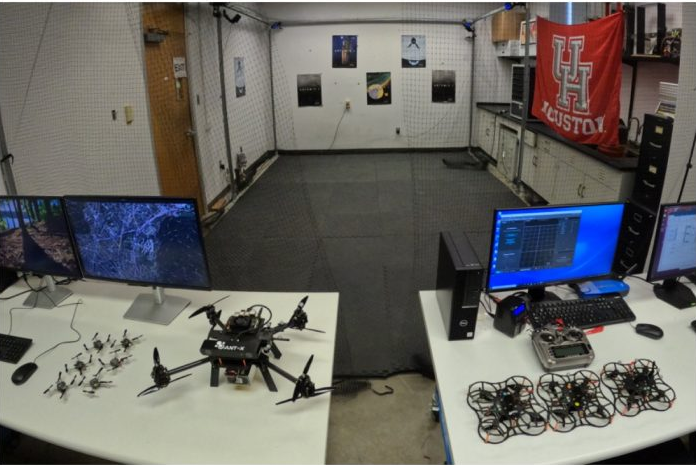}
    \caption{Advanced Learning, Artificial Intelligence, and Control Flight Arena at the University of Houston}
    \label{fig:FlightArena}
\end{figure}

\begin{figure}
    \centering
    \includegraphics[width = .49\textwidth]{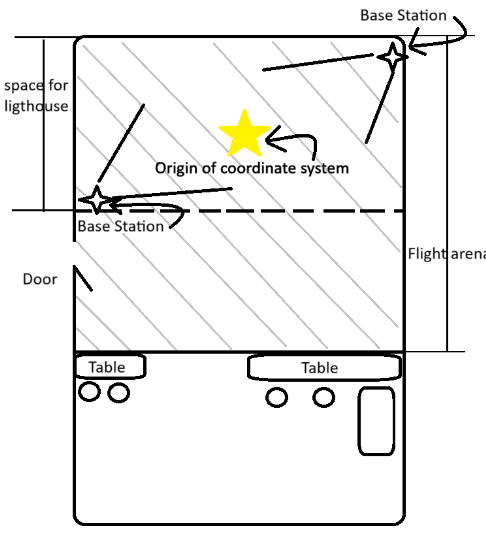}
    \caption{Diagram of Flight Arena}
    \label{fig:FlightArenaSketch}
\end{figure}

The implementation considered in this paper is scaled to and designed for a 4x3x2.5 [m]\footnote[1]{Note - The full lab space is 13x20x8 ft \cite{Cescon_Lab}, but the space available for use with the lighthouse positioning system is 13x10x8 ft*}
flight volume, illustrated in Figs.~\ref{fig:FlightArena}-\ref{fig:FlightArenaSketch}. This lab setup includes a positioning system with two diagonally opposed Bitcraze Lighthouse V2 Base Stations \cite{taffanel2021lighthouse}. These base stations operate by sending sweeps of infrared light that are observed by a Lighthouse deck on the Crazyflie. The Crazyflie uses the angles of these light sweeps to calculate its own position, which can then be communicated for use in a position/attitude controller via the Crazyradio system \cite{taffanel2021lighthouse}. %When working on this project, the intended expansion decks were the Lighthouse and Flow V2 decks. The Lighthouse deck is necessary for the Lighthouse positioning system to function, which allows for the observation of position in real time. The Flow deck uses distance and optical motion sensors pointed at the ground to facilitate more stable and consistent flight, but is not necessary for operations using the Lighthouse positioning system.

    \label{fig:BitcrazeCrazyflie21}
%\end{figure}
% \textcolor{red}{NOTE - details about the virtual machine, coding language, benefits and possibility of running natively, and so forth can be added but likely do not need to be covered given the scope of this paper.}

\subsection{Swarm Capabilities}
% \textcolor{red}{This section is new}
The Crazyflie platform has support for running swarms. The process for doing so is as follows:
\begin{enumerate}
    \item  The Uniform Resource Identifier (URI) (address) for communicating with each Crazyflie must be uniquely specified. 
    \item Then, each trajectory is defined, paired to a drone's URI, and the swarm will follow the trajectories. 
\end{enumerate}
   
%This framework makes testing trajectories on swarms of drones simple and straightforward.

\subsection{Waypoint Simulation Following on Crazyflies}
The Lighthouse positioning system allows for the use of a global coordinate system with the origin set to the ground at the center of the flight space, with base units for distance in millimeters \cite{taffanel2021lighthouse} \cite{CF_State} \cite{CF_Coords}. The trajectories and following waypoints are all made using a near identical global coordinate system with base units of meters. Scaling is performed to adjust the satellite trajectory to be flown on a Crazyflie.

% \textcolor{red}{Include details about running swarms}

%%%%%%%%%%%%%%%%%%%%%%%%%%%%%%%%%%%%%%%%%%%
%\section{Experimental Setup}
\section{Simulation Setup}
\subsection{Overview}
In this section, we present tests falling under two categories: spacecraft trajectory following  and neural network control output following. In the first category, natural motion and controlled spacecraft trajectories governed by the linearized CW equations in Hill's frame are calculated, formatted, scaled in time and distance,and followed by Crazyflies. Next, reinforcement learning-based neural network control systems trained for docking scenarios are used to generate closed loop control outputs that are scaled in distance and time and simulated by the Crazyflie drones. %, and multiple trajectories are tested simultaneously using swarms to show the swarm capabilities of the crazyflie platform.
%Note - can become four avenues if we include physical testing

\subsection{Testing Spacecraft Trajectory Following}
\begin{figure}[t!]%[hbt!]
    \centering
    \includegraphics[width = .49\textwidth]{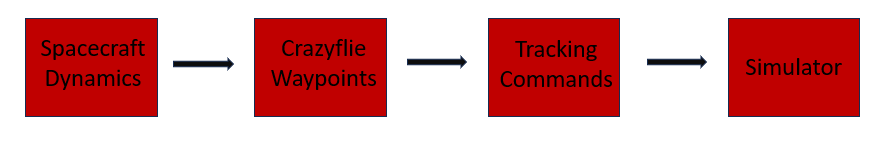}
    \caption{Workflow Block Diagram}
    \label{fig:WPBlockDiagram}
\end{figure}

The process of converting space dynamics to the terrestrial Crazyflie surrogates in simulation is depicted by the block diagram in Fig.\ref{fig:WPBlockDiagram}. First, the trajectory of the spacecraft under CW dynamics is calculated in the form of waypoints for spacecraft to follow. Second, this spacecraft trajectory is scaled in distance to an appropriate size to be flown on a Crazyflie 2.1 within the laboratory space and the duration is shortened to represent the entire path well within the battery life of the Crazyflie. Third, the gym environment is initialized and the waypoints are sent to the Crazyflie. Crazyflie tracking commands are generated using the PID controller introduced in Sec.~\ref{Sec:background}\cite{mellinger2011minimum}. Fourth, a Crazyflie flight following these waypoints is simulated using the PID control commands. 
These steps are described in more detail in the following subsections.

\subsubsection{Space Dynamics Simulator}
Before being simulated on the Crazyflie surrogate, a spacecraft's motion in the space-scale Hill's frame is determined. The continuous form of the CW equations introduced in Eq.~\eqref{eqn: CWEquation}, were solved using a Runge-Kutta 45 solver. Note that the output is formatted such that the orbital period and simulation runtime are adjustable independently from each other. The spacecraft dynamics simulators were created as waypoint generating functions which convert the space-scale outputs of the Clohessy-Wiltshire solutions to the drone-scale and integrates into the existing trajectory input tools available in \texttt{gym-pybullet-drones}. 

\subsubsection{Waypoint Generation}
\begin{figure}[b!]%[htb!]
    \centering
    \includegraphics[width = .49\textwidth]{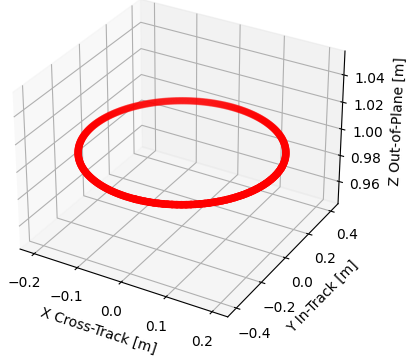}
    \caption{Lab-Scale, In-Plane Natural Motion Trajectory}
    \label{fig:NMTTrajectory}
\end{figure}

Depending on the time span and the solver step size chosen, the simulator generates a number of waypoints for the simulated drone to follow. Before the waypoints are sent to the simulator, the distance is scaled from the space-scale to the lab-scale, where the lab space is assumed to not be any larger than 4x3x2.5 meters. The distance scale is a factor of 4e3. The waypoints generated serve as discrete position inputs for the drone simulation system, where the number of waypoints solved correlates to the control frequency multiplied by the chosen time period. Fig.\ref{fig:NMTTrajectory} shows an example of the waypoints generated for an in-plane NMT where both the chief and the target remain in the same relative orbital plane (the $z$-axis, in this and future examples).
The type of trajectory generated by the Clohessey-Wiltshire equations are dependent on the initial states of the system. The initial conditions used in this example are as follows:% (note distances are in meters and velocities are in meters per second),
\begin{equation}
    \centering
    \begin{matrix}
        x =  800  & \dot{x} = 0.16 \\
        \\
        y = \frac{2\dot{x}}{n} & \dot{y} = -2nx \\
        \\
        z = 0 & \dot{z} = 0\\
    \end{matrix}
    \label{eqn: ICs}
\end{equation}
with $x,y,z$ [m] and $\dot{x}, \dot{y}, \dot{z}$ [m/s].
These initial conditions were arbitrarily chosen to loosely model in-space/on-orbit servicing, assembly, and manufacturing (OSAM/ISAM) servicing between two bodies \cite{arney2021orbit}. This simplified simulator was used to generate NMTs and completely makes up the 'Spacecraft Dynamics' block in Fig.\ref{fig:WPBlockDiagram}.

 \subsubsection{Tracking Commands}
The drone's initial position is chosen to equal the lab-scale initial positions in the space dynamics simulator with the starting rotational position $\theta_0 =  \begin{bmatrix} 0, 0, \pi/2\end{bmatrix}^T$ relative to the origin. This initial rotational position starts the drone with all four propellers facing upwards. The simulated drone acts as the deputy spacecraft surrogate and the origin of the inertially fixed frame as the chief. The generated waypoints are then injected into the flight simulator where the drone is controlled by the PID controller introduced in Section 2. 
 
\subsection{Space Dynamics Through Neural Network}
\begin{figure}[htb!]
    \centering
    \includegraphics[width = .49\textwidth]{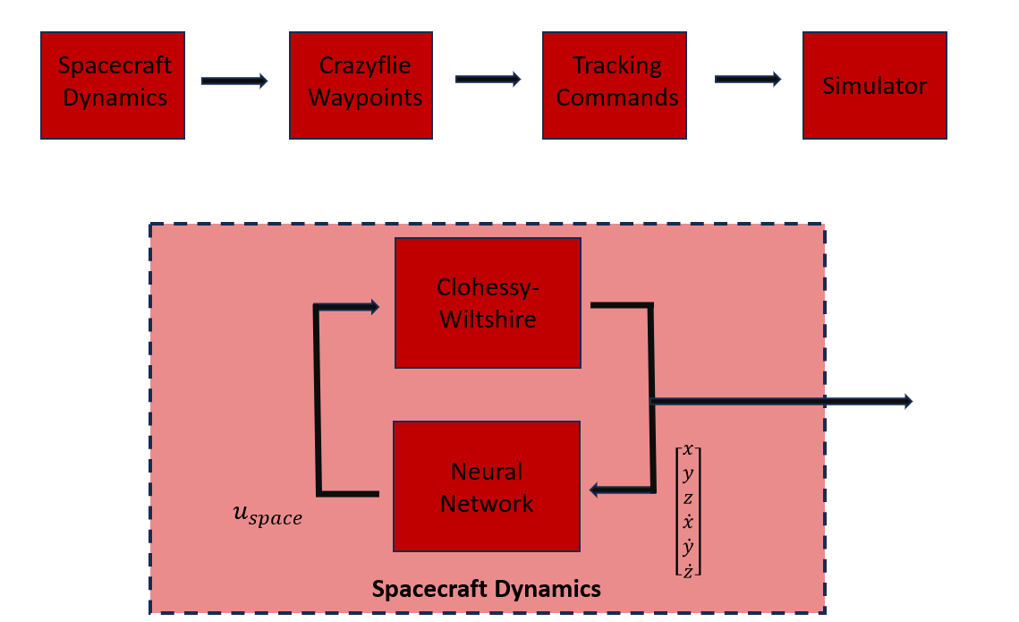}
    \caption{Workflow Block Diagram with Neural Network}
    \label{fig:ComBlockDiagram}
\end{figure}

An alternative method of generating a trajectory in the space-scale is through the use of a neural network control system trained with reinforcement learning for specific on-orbit scenarios (e.g. docking and inspection) provided by the work outlined in \cite{hamilton2022ablation}. In this implementation, the trained neural network outputs control the thrusts to the simulation with linearized Clohessy-Wiltshire equations that govern the satellite motion. This neural network controlled trajectory is then formatted and scaled to be flown on Crazyflies, the process for which is shown in Figure \ref{fig:ComBlockDiagram}.

%\subsection{Swarm Surrogates}
%Testing with swarms is of great relevance for a platform used to test satellite dynamics teresstrially. This is because a high value use of testing spacecraft trajectories is in risk assessment. As such, the trajectories made and tested in the CWH and neural network tests (NMTs and docking scenarios) have been taken and fed as waypoints to multiple crazyflies simultaneously. First, two spacecraft docking the same target satellite from different starting points is simulated, and then a scenario is run wherein one surrogate docks while another is following it in a natural motion trajectory.

%%%%%%%%%%%%%%%%%%%%%%%%%%%%%%%%%%%%%%%%%%%
\section{Results}
\begin{figure}[b]%[htb!]
    \centering
    \includegraphics[width = .49\textwidth]{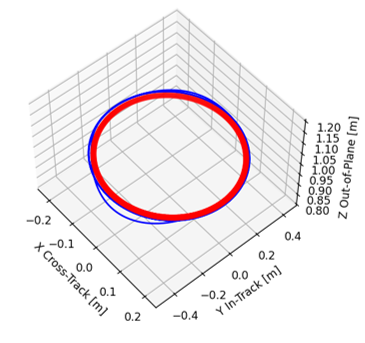}
    \caption{3D Plot of In-Plane Waypoint Tracking with Tuned PID}
    \label{fig:InPlaneTuned}
\end{figure}

\begin{figure}[htb!]
    \centering
    \begin{subfigure}[b]{0.5\textwidth}
        \centering
        \includegraphics[width =\textwidth]{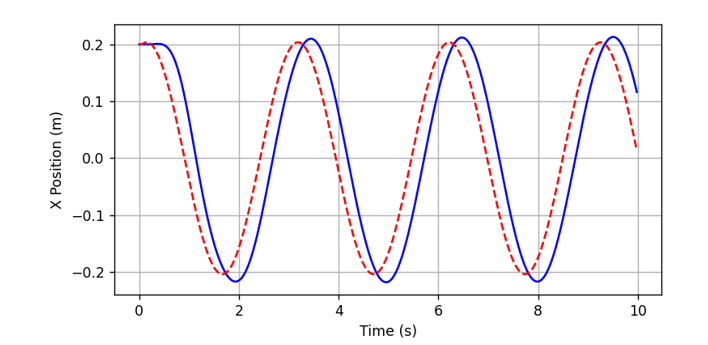}
        \caption{}
        \label{fig:2DWPTracking1}
    \end{subfigure}
    \hfill
    \begin{subfigure}[b]{0.5\textwidth}
        \centering
        \includegraphics[width =.98\textwidth]{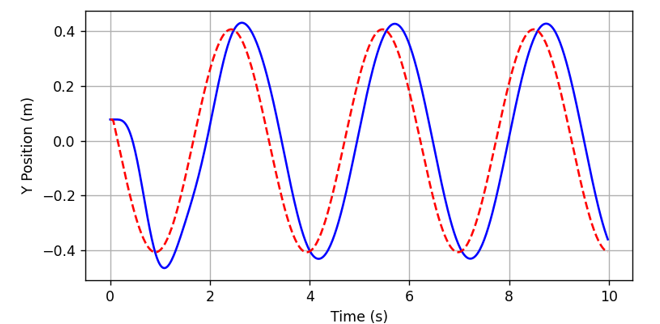}
        \caption{}
        \label{fig:2DWPTracking2}
    \end{subfigure}
    \hfill
    \caption{Drone X (a) and Y (b)  Positions (Blue) Compared to Waypoints (Red)}
    \label{fig:2DWPTracking}
\end{figure}

Figures \ref{fig:InPlaneTuned}-\ref{fig:2DWPTracking} show results of the waypoint trajectory tracking with the PID controller performed on an in-plane NMT in the gym environment. This simulated experiment was performed over the span of three complete orbital revolutions during a time span of ten seconds, with initial conditions previously introduced in Eq. \ref{eqn: ICs}.
%Waypoint trajectory tracking through the PID controller was performed on an in-plane NMT in the gym environment, the results of which can be seen in Figure \ref{fig:InPlaneTuned} and Figure \ref{fig:2DWPTracking}. 
%The initial conditions for this experiment were the same as previously introduced in Eq. \ref{eqn: ICs} and was performed over the span of three complete orbital revolutions over a time span of ten seconds. It should be noted that the orbital frequency is a tunable parameter and can be chosen by the simulation designer. 

\begin{figure}[htb!]
    \centering
    \includegraphics[width = .49\textwidth]{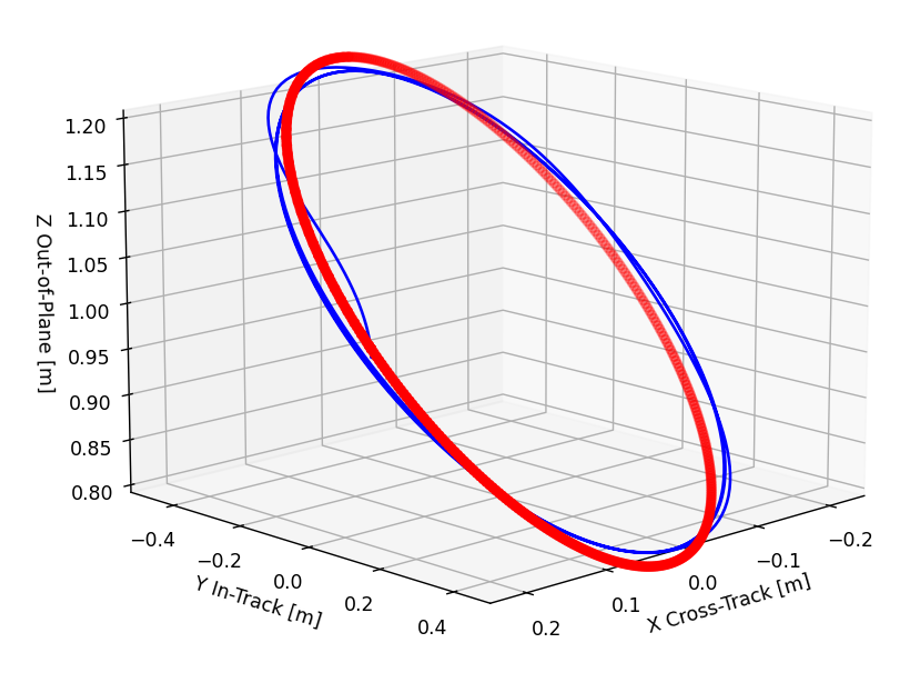}
    \caption{3D Plot of Out-of-Plane Waypoint Tracking with Tuned PID}
    \label{fig:OutPlaneTuned}
\end{figure}

\begin{figure}[htb!]
    \centering
    \begin{subfigure}[b]{0.5\textwidth}
        \centering
        \includegraphics[width =\textwidth]{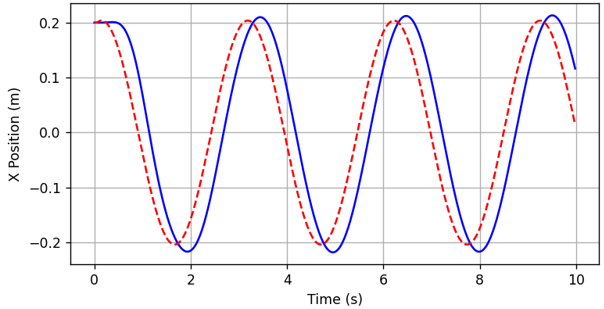}
        \caption{}
        \label{fig:2DWPTracking1}
    \end{subfigure}
    \hfill
    \begin{subfigure}[b]{0.5\textwidth}
        \centering
        \includegraphics[width =\textwidth]{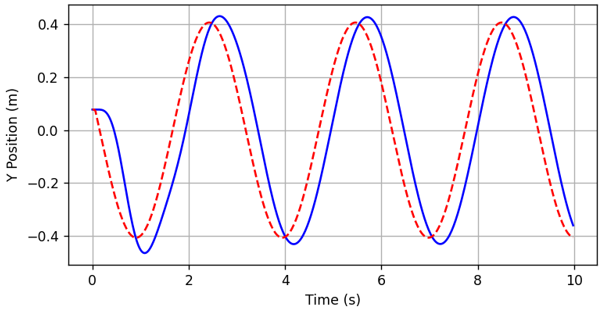}
        \caption{}
        \label{fig:2DWPTracking2}
    \end{subfigure}
    \hfill
    \begin{subfigure}[b]{0.5\textwidth}
        \centering
        \includegraphics[width =\textwidth]{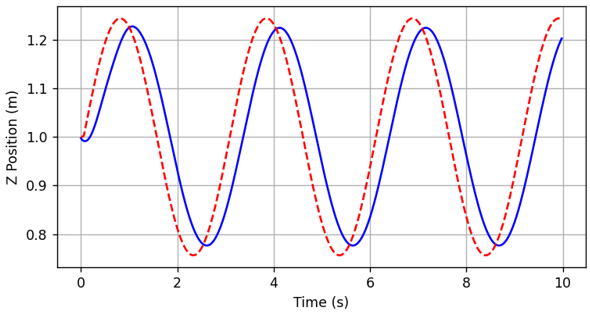}
        \caption{}
        \label{fig:2DWPTracking3}
    \end{subfigure}
    \caption{Drone X (a), Y (b), and Z (c)  Positions (Blue) Compared to Waypoints (Red) for Out-of-Plane NMT}
    \label{fig:OutPlaneTuned2}
\end{figure}

A more general case of a relative elliptical orbit is that of out-of-plane trajectories where the follower and target satellite orbit in different orbital planes relative to the earth centered frame. In order to create an out of plane elliptical NMT, the position and velocity initial conditions in the $x$ and $y$ axis remained the same as the previous case with the position and velocity in the $z$ axis becoming non-zero. Arbitrary values were assigned to the states in the $z$ axis,
\begin{equation}
    \centering
    \begin{matrix}
            z = 1 & \dot{z} = 1\\
    \end{matrix}
\end{equation}

and the NMT generated was simulated in the gym environment. Results of this simulated experiment are given in Figs. \ref{fig:OutPlaneTuned}-\ref{fig:OutPlaneTuned2}.

% Note - The ability to create controlled trajectories is explained and the results of such a flight is shown in the appendix 

As explained in the previous section, the origin is representative of a target satellite and the trajectory shown is the chaser's trajectory relative to the target. 

\subsection{Crazyflie Tracking Reinforcement Learning Outputs}
\begin{figure}[!h]
    \centering
    \includegraphics[width = .49\textwidth]{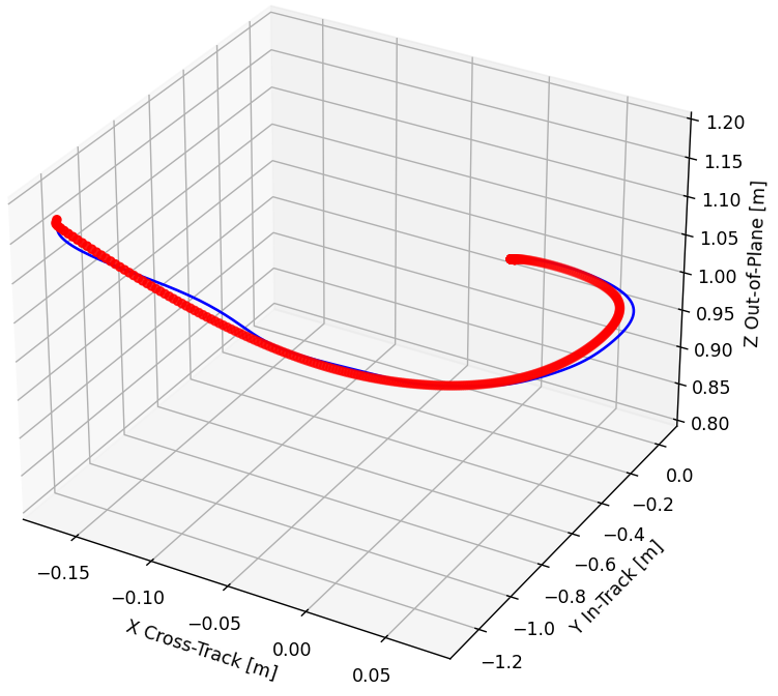}
    \caption{3D Plot of RL-Trained Docking Simulation, Target Spacecraft is Assumed to be at coordinates [0,0,1]}
    \label{fig:RLTracking1}
\end{figure}

\begin{figure}[!h]
    \centering
    \begin{subfigure}[b]{0.5\textwidth}
        \centering
        \includegraphics[width =\textwidth]{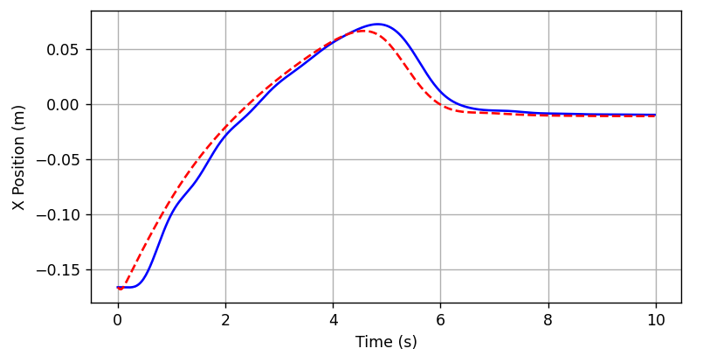}
        \caption{}
        \label{fig:2DWPTracking1}
    \end{subfigure}
    \hfill
    \begin{subfigure}[b]{0.5\textwidth}
        \centering
        \includegraphics[width =\textwidth]{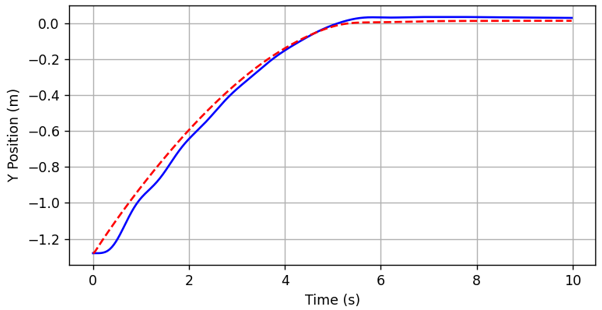}
        \caption{}
        \label{fig:2DWPTracking2}
    \end{subfigure}
    \hfill
    \begin{subfigure}[b]{0.5\textwidth}
        \centering
        \includegraphics[width =\textwidth]{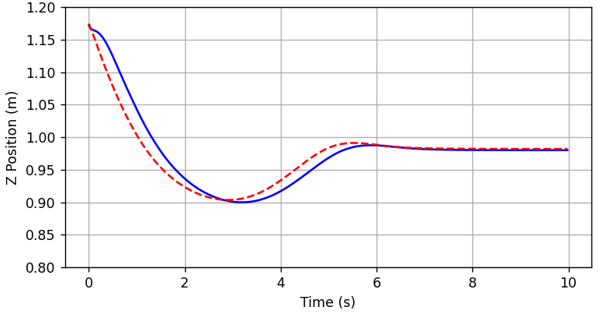}
        \caption{}
        \label{fig:2DWPTracking3}
    \end{subfigure}
    \caption{Drone X (a), Y (b), and Z (c)  Positions (Blue) Compared to Waypoints (Red) for RL-Trained Docking Simulation}
    \label{fig:RLTracking2}
\end{figure}

A trained reinforcement learning neural network \cite{ravaioli2022safe} for a spacecraft docking scenario was inserted into the control scheme, as depicted in Figure \ref{fig:ComBlockDiagram}, where the control thrusts from the trained neural network controller were fed into the Clohessy-Wiltshire linearized model, Eq. \ref{eqn: CWEquation}. A pre-generated trajectory for a docking scenario from this neural network was saved as a dictionary and was used as the input trajectory in the gym environment. In this scenario, a chaser spacecraft attempts to dock with a target spacecraft (located at the origin in the non-inertial Hill's frame) within a given time horizon. A simulation time horizon of 10 seconds was chosen and the docking scenario was simulated in the gym environment, the results of which can be seen in Figure \ref{fig:RLTracking1} and Figure \ref{fig:RLTracking2}.  

\begin{figure}[!h]
    \centering
    \includegraphics[width = .49\textwidth]{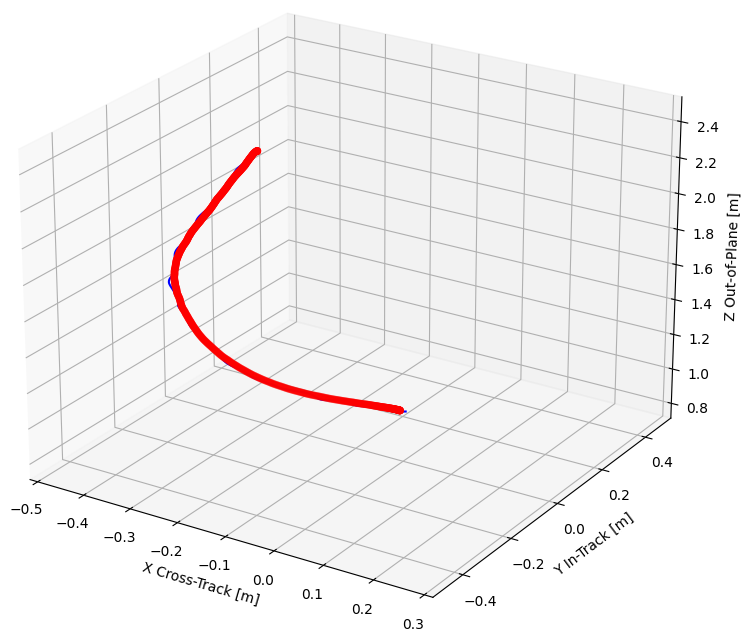}
    \caption{3D Plot of Integrated RL Docking Simulation, Target Spacecraft is Assumed to be at coordinates [0,0,1]}
    \label{fig:IRL1}
\end{figure}

\begin{figure}[!h]
    \centering
    \begin{subfigure}[b]{0.5\textwidth}
        \centering
        \includegraphics[width =\textwidth]{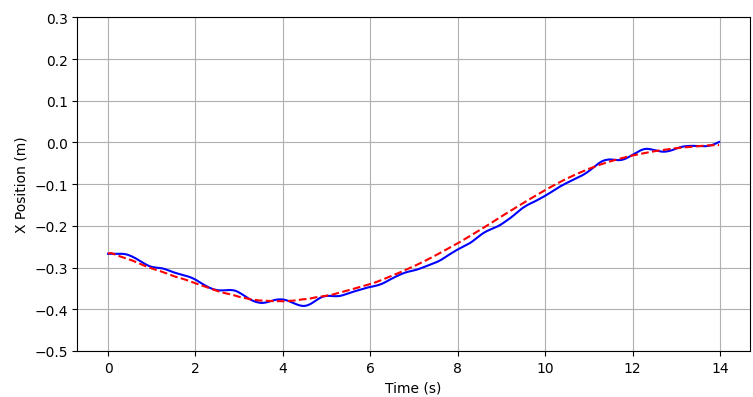}
        \caption{}
        \label{fig:IRL21}
    \end{subfigure}
    \hfill
    \begin{subfigure}[b]{0.5\textwidth}
        \centering
        \includegraphics[width =\textwidth]{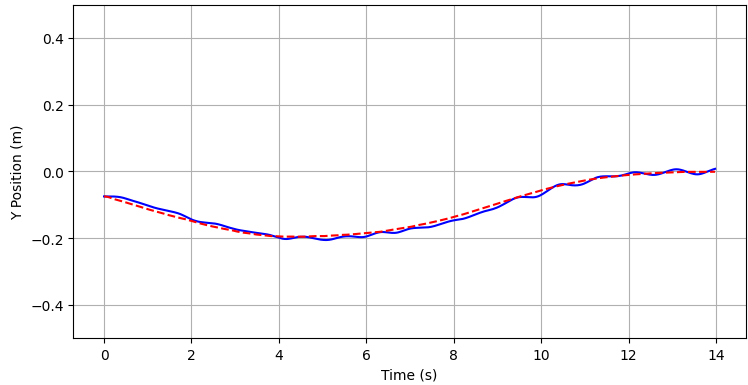}
        \caption{}
        \label{fig:IRL22}
    \end{subfigure}
    \hfill
    \begin{subfigure}[b]{0.5\textwidth}
        \centering
        \includegraphics[width =\textwidth]{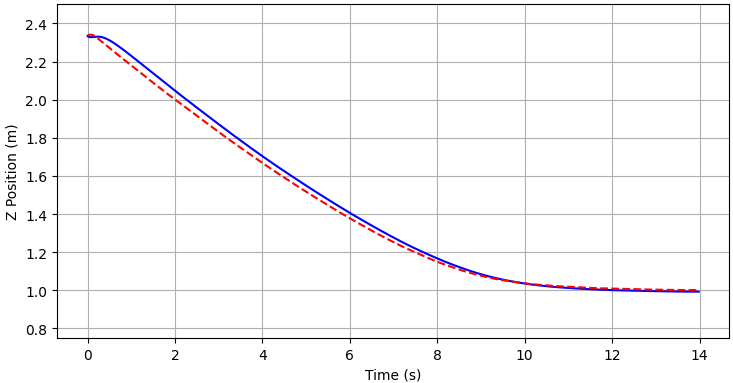}
        \caption{}
        \label{fig:IRL23}
    \end{subfigure}
    \caption{Drone X (a), Y (b), and Z (c)  Positions (Blue) Compared to Waypoints (Red) for Integrated RL Docking Simulation}
    \label{fig:IRL2}
\end{figure}

A different neural network was trained for this same scenario and, instead of the trajectory generated being externally saved, the neural network control scheme was fully integrated with the \texttt{gym-pybullet-drones} simulator. The neural network can be trained with assigned values for the chief initial conditions. For this network, the initial conditions for the chief were chosen to all equal zero. Once trained, the neural network can control the deputy satellite to dock from a range of initial conditions chosen by a random number generator (RNG) with a seeded value. The initializer used for this scenario includes a built-in safety feature which ensures that the initial velocity of the deputy spacecraft does not violate a set maximum velocity safety limit (there exits a limit in which docking cannot be conducted safely). Figure \ref{fig:IRL1} and Figure \ref{fig:IRL2} show an implementation with a unique set of initial conditions. Note that the time horizon for this example was expanded to 14 seconds.

% \subsection{Swarm Surrogates}
% The trajectories created heretofore have been taken and now they will be implemented in swarms of drones to visualize certain safety constraints within relative motion scenarios as well as to display the ability of this platform in testing with swarms. First, \textcolor{red}{Two drones are flown, both docking from different initial conditions}. This is done to showcase a high risk situation involving a potential crash, which does happen in our results shown. It is worth noting that the results shown have not implemented run time assurance or any other such method to guarantee safety, which is why the drones \textcolor{red}{crash/do what they do}. The implementation of a method to guarantee safety, such as run time assurance, is a high value focus for future work.

% \textcolor{red}{insert figure here}

% Now the results of a swarm wherein one drone is flying in an ellipse and another drone is docking will be shown and discussed. The motivations in selecting this situation remain the same as for the previous swarm experiment, but this time the trajectories for the surrogate following natural motion and the surrogate docking do not necessarily need to cross. A high value usecase for testing relative satellite motion terrestrially is to ass the risk of different trajectories, and this is an example of a scenario wherein a potential collision can be avoided!

% \textcolor{red}{insert figure here}

%%%%%%%%%%%%%%%%%%%%%%%%%%%%%%%%%%%%%%%%%%%
\section{Conclusions}
The capabilities of the Bitcraze's Crazyflie platform in testing satellite autonomy terrestrially were demonstrated in simulation. Simple relative satellite motion governed by the linearized Clohessy-Wiltshire equations in Hill's frame were fully implemented for testing simple in-plane and out-of-plane natural motion trajectories. Additionally, two tests of reinforcement learning controlled docking scenarios were successfully performed. The first RL agent's test was ran by saving the output of the RL agent's controlled flight, formatting it to be flown on the Crazyflies, then simulated. The second RL agent calculated the trajectory which was then directly formatted and flown by the Crazyflie in the gym environment.
% Finally, the trajectories have been flown on swarms to show the capabilities in displaying relative spacecraft dynamics, as well as visualizing a certain extent of desirable safety constraints.

\subsection{Future Work}
The next step in this research is translating the work completed in the \texttt{gym-pybullet-drones} simulation space to physical flight arena using actual Crazyflie drones. It is anticipated that a challenge in this transfer is the Sim2Real Correlation Coefficient (SRCC), a measure of how well a simulation matches real-world performance, which has not been extensively tested for \texttt{gym-pybullet-drones}. Imperfect models of aerodynamic effects, latency, and actuator saturation all impact this coefficient and may be reviewed. Additionally, the base PID controller can be further tuned to reduce any tracking error encountered in the simulation. When successfully physically implemented, this research may have a variety of use cases including testing novel inspection techniques of uncooperative target satellites \cite{van2023deep} and cubesat proximity operations \cite{bowen2015cubesat}. Visualizing physical performance of these types of missions can aid in evaluation and serve as a foundation for tuning.

%%%%%%%%%%%%%%%%%%%%%%%%%%%%%%%%%%%%%%%%%%%
\acknowledgments
The authors thank Nathaniel Hamilton, PhD and James Cunningham for providing their support and expertise. This work is supported by the Department of Defense (DoD) through the Air Force Research Laboratory (AFRL) Minority Leaders Research Collaboration Program (ML-RCP). The views expressed are those of the authors and do not reflect the official guidance or position of the United States Government, the Department of Defense or of the United States Air Force.

%%%%%%%%%%%%%%%%%%%%%%%%%%%%%%%%%%%%%%%%%%%
\bibliographystyle{IEEEtran}
\bibliography{references}

%%%%%%%%%%%%%%%%%%%%%%%%%%%%%%%%%%%%%%%%%%%
% \thebiography

% % This biostyle allows you to insert your photo size 1in X 1.25in

\begin{biographywithpic}{Arturo de la Barcena III}{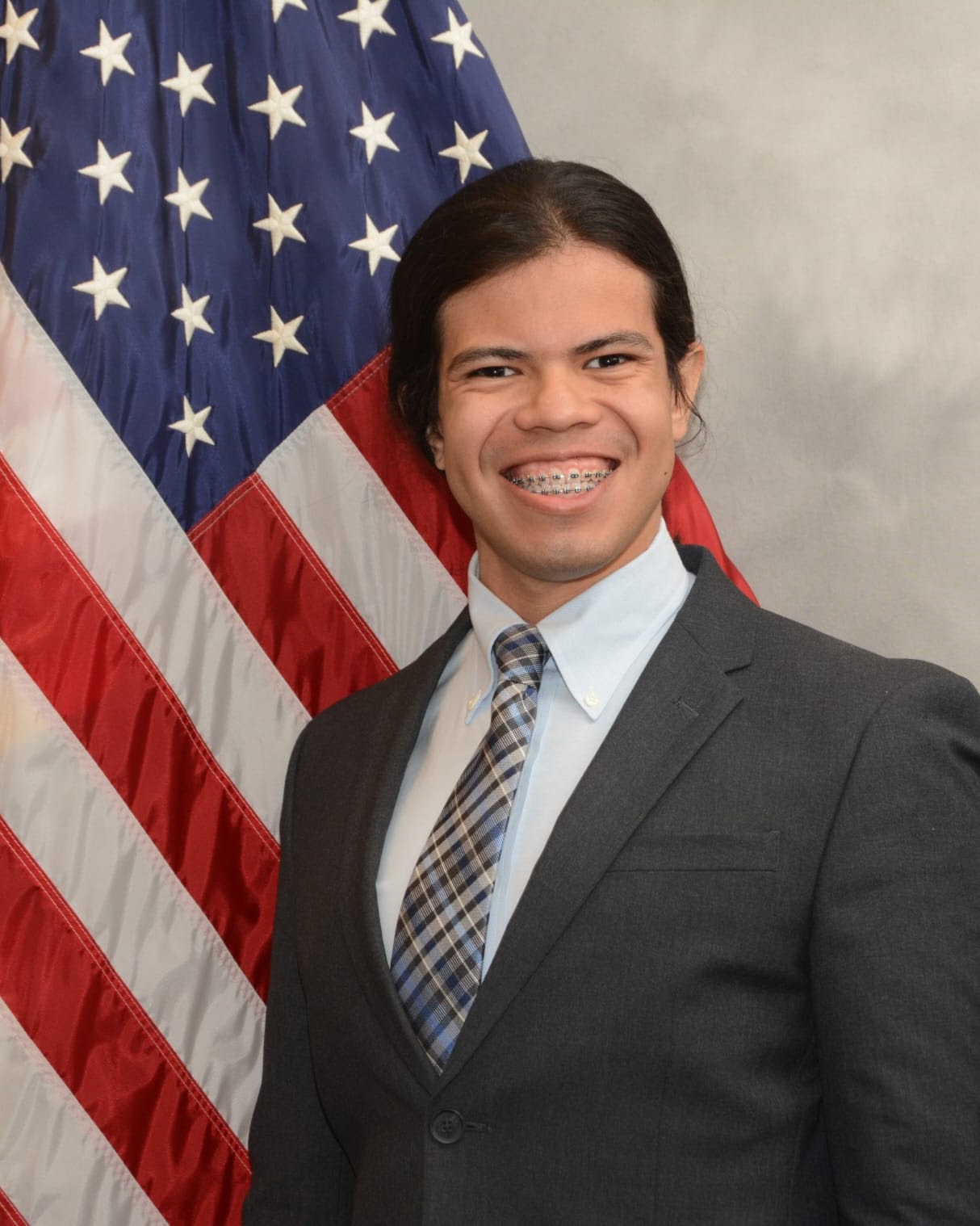}
is an intern working under Dr. Kerianne Hobbs and a mechanical engineering master's student at the University of Houston. His primary research interests are in air and space vehicle control and autonomy and has been involved with the Advanced Learning, Artificial Intelligence and Control lab at the University of Houston for over two years.
\end{biographywithpic}

\begin{biographywithpic}{Collin J. Rhodes}{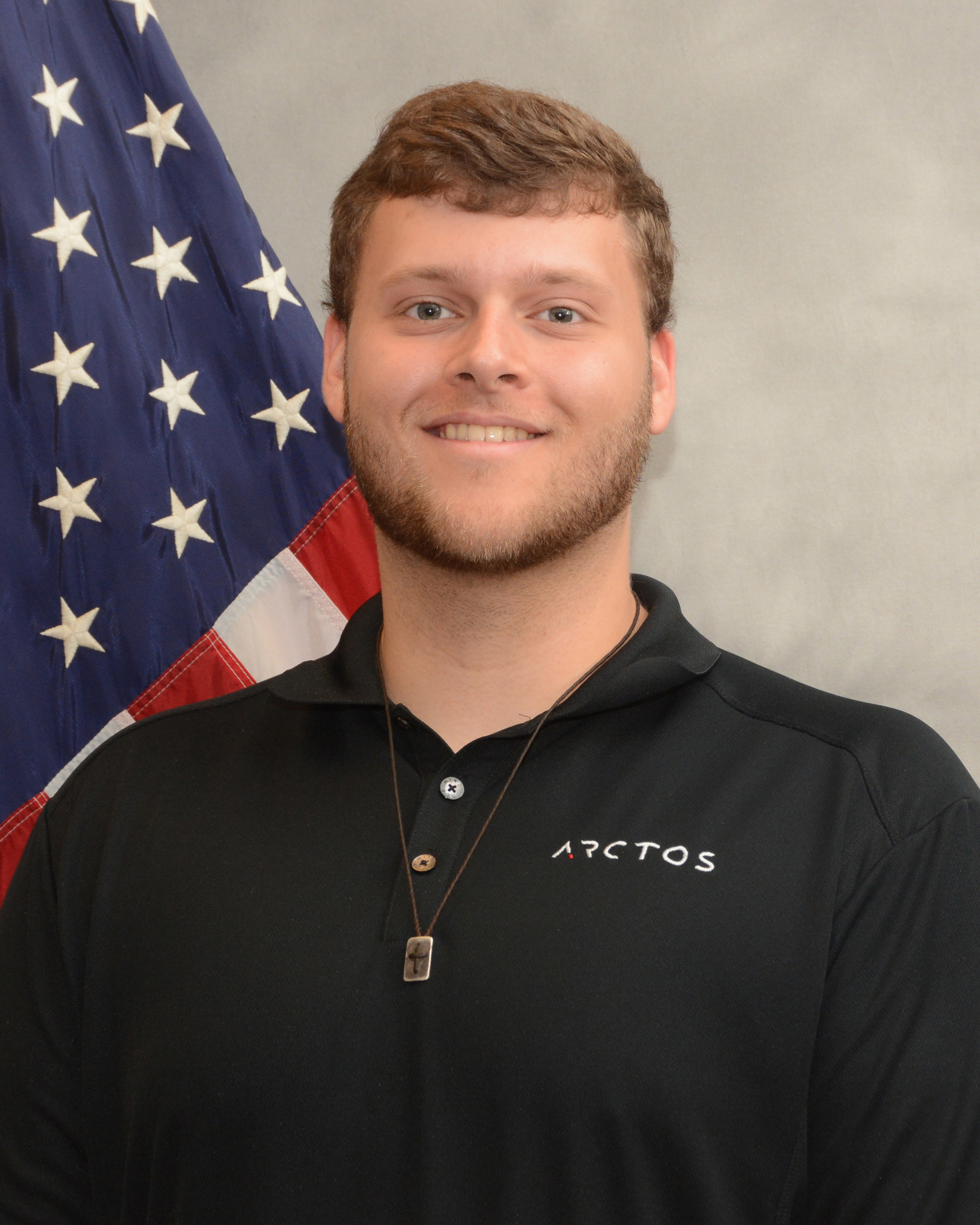}
is a contracted intern for the Air Force Research Lab working directly under Dr. Hobbs. He is a graduate student studying Mechanical Engineering with a focus in controls at the University of Houston. In addition to research and classes, Collin is also the President of Cougar Racing, a student organization wherein students compete against other universities in designing and building a race car every year.
\end{biographywithpic}

\begin{biographywithpic}{John McCarroll}{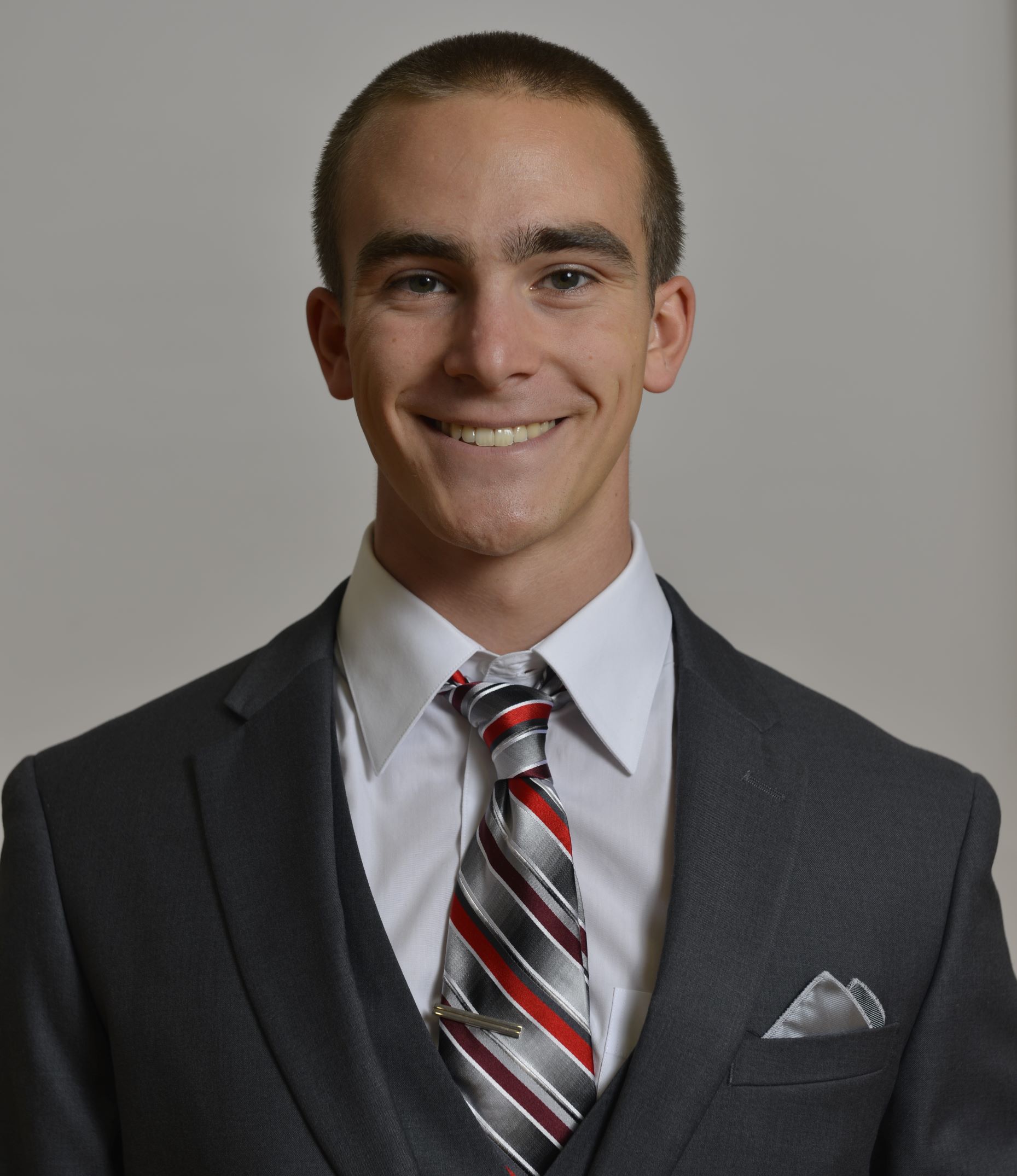} is a Junior Autonomy Engineer on the Autonomy Capability Team (ACT3) at the Air Force Research Laboratory. He works on the Safe Autonomy team, supporting development efforts and experimenting with reinforcement learning. John received his BS degree in Software Engineering from Rochester Institute of Technology, where he gained experience in neural architecture search using genetic algorithms, CNNs applied to games and art, data warehousing, and data visualization.
\end{biographywithpic} 

\begin{biographywithpic}{Marzia Cescon}{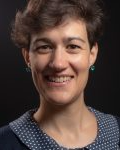}
is the David C. Zimmerman Assistant Professor of Mechanical Engineering at the University of Houston, Houston, TX. She is the founder and director of the Advanced Learning, Artificial Intelligence and Control laboratory, a multidisciplinary effort developing novel computational methods and tools for learning-based decision making and control of complex and unknown dynamical systems. Her interests span biomedical control systems, multi-agent systems, flight control and aerospace control, and safe autonomy. Dr. Cescon earned a Bachelor Degree in Information Engineering and a Master Degree in Control Systems Engineering both from the University of Padua, Italy, and received the Ph.D. degree in Automatic Control from the Automatic Control Department at Lund University, Sweden. She has held several research positions including at the University of California, Santa Barbara, the Melbourne School of Engineering at University of Melbourne, and the Harvard John A. Paulson School of Engineering and Applied Sciences at Harvard University. 

\end{biographywithpic}

\begin{biographywithpic}{Kerianne L. Hobbs}{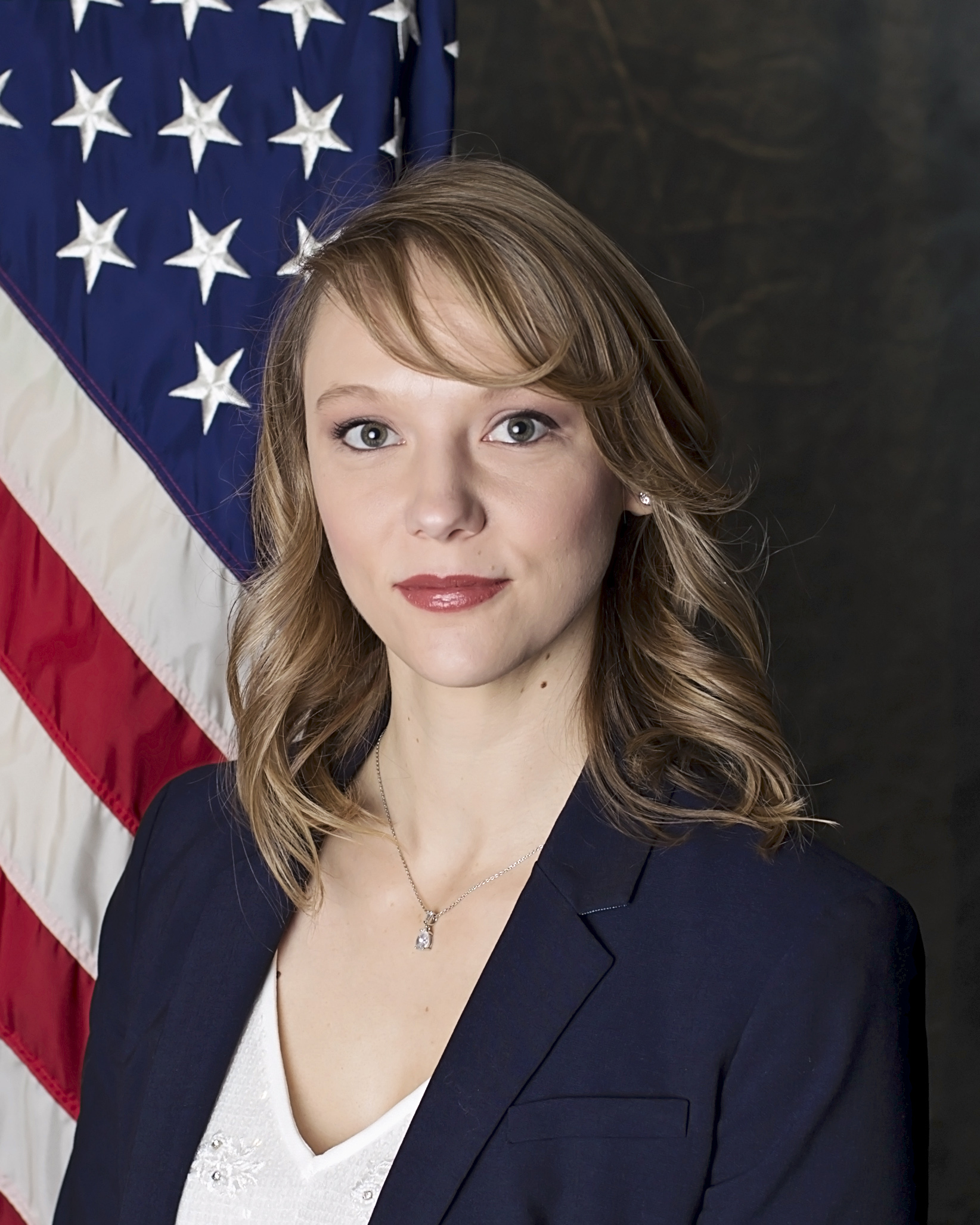}
is the Safe Autonomy and Space Lead on the Autonomy Capability Team (ACT3) at the Air Force Research Laboratory. There she investigates rigorous specification, analysis, bounding, and intervention techniques to enable safe, trusted, ethical, and certifiable autonomous and learning controllers for aircraft and spacecraft applications. Her previous experience includes work in automatic collision avoidance and autonomy verification and validation research. Dr. Hobbs was selected for the 2020 AFCEA 40 Under 40 award and was a member of the team that won the 2018 Collier Trophy (Automatic Ground Collision Avoidance System Team), as well as numerous AFRL Awards.  Dr. Hobbs has a BS in Aerospace Engineering from Embry-Riddle Aeronautical University, an MS in Astronautical Engineering from the Air Force Institute of Technology, and a Ph.D. in Aerospace Engineering from the Georgia Institute of Technology.  \end{biographywithpic}
\end{document}